\documentclass[runningheads]{llncs}
\usepackage{graphicx}
\usepackage{subcaption}
\usepackage{hyperref}
\usepackage{comment}
\usepackage{multirow}
\usepackage{booktabs}
\usepackage{listings}
\usepackage{mdframed}
\usepackage{xcolor}
\usepackage{tabularx}
\usepackage{eurosym}

\definecolor{codegreen}{rgb}{0,0.6,0}
\definecolor{codegray}{rgb}{0.5,0.5,0.5}
\definecolor{codepurple}{rgb}{0.58,0,0.82}
\definecolor{backcolour}{rgb}{0.95,0.95,0.92}

\lstdefinestyle{mystyle}{
    backgroundcolor=\color{backcolour},   
    commentstyle=\color{codegreen},
    keywordstyle=\color{magenta},
    numberstyle=\tiny\color{codegray},
    stringstyle=\color{codepurple},
    basicstyle=\ttfamily\footnotesize,
    breakatwhitespace=false,         
    breaklines=true,                 
    captionpos=b,                    
    keepspaces=true,                 
    numbers=left,                    
    numbersep=5pt,                  
    showspaces=false,                
    showstringspaces=false,
    showtabs=false,                  
    tabsize=2
}

\newcommand{\cameraready}[1]{#1}

\begin{document}
\title{Do LLMs Really Adapt to Domains?\\An Ontology Learning Perspective}
\titlerunning{Do LLMs Really Adapt to Domains? An Ontology Learning Perspective}

\author{Huu Tan Mai\inst{1,2}\orcidID{0009-0003-6584-4212} \and
Cuong Xuan Chu\inst{2} \and
Heiko Paulheim\inst{1}\orcidID{0000-0003-4386-8195}}
\authorrunning{Mai et al.}
\institute{University of Mannheim, Data and Web Science Group, 68161 Mannheim, Germany \and
Bosch Center for AI, 71272 Renningen, Germany\\
\email{\{huutan.mai,cuongxuan.chu\}@de.bosch.com}\\
\email{\{huu.tan.mai,heiko.paulheim\}@uni-mannheim.de}}
\maketitle              % typeset the header of the contribution
\begin{abstract}
Large Language Models (LLMs) have demonstrated unprecedented prowess across various natural language processing tasks in various application domains. Recent studies show that LLMs can be leveraged to perform lexical semantic tasks, such as Knowledge Base Completion (KBC) or Ontology Learning (OL). However, it has not effectively been verified whether their success is due to their ability to reason over unstructured or semi-structured data, or their effective learning of linguistic patterns and senses alone. This unresolved question is particularly crucial when dealing with domain-specific data, where the lexical senses and their meaning can completely differ from what a LLM has learned during its training stage. This paper investigates the following question: \textit{Do LLMs really adapt to domains and remain consistent in the extraction of structured knowledge, or do they only learn lexical senses instead of reasoning?} To answer this question and, we devise a controlled experiment setup that uses WordNet to synthesize parallel corpora, with English and gibberish terms. We examine the differences in the outputs of LLMs for each corpus in two OL tasks: relation extraction and taxonomy discovery. Empirical results show that, while adapting to the gibberish corpora, off-the-shelf LLMs do not consistently reason over semantic relationships between concepts, and instead leverage senses and their frame. However, fine-tuning improves the performance of LLMs on lexical semantic tasks even when the domain-specific terms are arbitrary and unseen during pre-training, hinting at the applicability of pre-trained LLMs for OL.
\end{abstract}
\section{Introduction}

Knowledge Bases (KB) and ontologies play a key role in structuring and organizing knowledge across domains, and offer powerful solutions to link data that would otherwise remain unstructured (such as text). As of now, many sources of data can be used as ontologies, of varying specificity. For instance, WordNet~\cite{wordnet}, ConceptNet~\cite{conceptnet} and WebIsA~\cite{webisadb} contain common knowledge, whereas KBs such as Unified Medical Language System (UMLS)~\cite{Bodenreider2004TheUM} and GeoNames~\cite{geonames} have their own domain specificities (respectively, medical and geographic). For knowledge-intensive applications, access to structured data is of utmost importance, but creating it is a very tedious and time-consuming process that inevitably demands domain expertise. Ontology Learning (OL) is a field of artificial intelligence that concerns itself with automatically identifying terms, types and axioms between them from unstructured or structured information sources such as text or KBs. \cameraready{In particular, OL deals with identifying hyponymy (resp. hypernymy) relations in a KB. That is, for pairs of concepts $(A, B)$ in the KB, one wants to infer whether or not Concept $A$ is a subclass (resp. a superclass) of Concept $B$.}

In the past few years, Large Language Models (LLMs), such as GPT-3~\cite{brown2020language}, GPT-4, LLaMa2~\cite{touvron2023llama} or Falcon-40B~\cite{falcon40b} have displayed unprecedented prowess in many NLP applications across various domains. LLMs are language models with a very large parameter count that are trained with enormous amounts of textual data. Hence, they are equipped with common knowledge and have shown remarkable success in generating text. Furthermore, LLMs have made it possible to capture the meaning of text and reason about it, providing a promising alternative for knowledge-intensive tasks such as KB completion and ontology-related tasks like OL and OM (Ontology Matching). Recent studies have shown that LLMs could be viewed as Knowledge Bases~\cite{petroni2019language}, storing knowledge incorporated in their parameters (e.g. factual knowledge~\cite{sun2023headtotail}, event knowledge~\cite{kauf2023event}, commonsense knowledge~\cite{commonsense,zhao2023commonsense}, etc). In particular, LLMs4OL~\cite{giglou2023llms4ol} introduces a novel approach to OL using LLMs: the work provides comprehensive empirical evidence that, although requiring fine-tuning for better performance, LLMs can work as effective ontology learners on specialized datasets. 

However, there are several challenges that come with LLMs, and that have been left relatively unexplored in the frame of ontology learning. They include: (1) LLMs are susceptible to hallucinate~\cite{li2023halueval}, i.e. provide generate text that is syntactically sound but factually incorrect. (2) LLMs are trained on massive corpora of textual data and acquire common knowledge, but their few - or zero - shot generalizability and adaptability to unknown domains remains relatively undiscussed. Studying these two
aspects is crucial to better understand the current limits of LLMs, and is ever so needed since they are being become increasingly prominent for domain-specific uses. On the one hand, it is effectively possible to get LLMs to \textit{adapt} to a domain by fine-tuning, but such a process requires labeled data in this domain and can be computationally expensive. On the other hand, \textit{generalizability} is also an extremely valuable quality for OL in domain-specific settings. To illustrate this, consider the Examples~\ref{ex_real_1} and ~\ref{ex_gib_1}.
\begin{figure}[t!]
\begin{mdframed}
\begin{example}[macaron]\label{ex_real_1}
    a sweet meringue-based confection made with egg white, icing sugar, granulated sugar, almond powder or ground almond, and food colouring.
\end{example}
\begin{example}[twiglomptoroa]\label{ex_gib_1}
    a becsverdecoroa meringue-based becsverdecoroal made with egg white, icing shifbousadu stintsobroa, granulated shifbousadu stintsobroa, almond powder or ground almond, and food colouring.
\end{example}
\end{mdframed}
\end{figure}

In OL, a LLM may be able to identify that a \textit{macaron} is a subclass of \textit{confection} that is made from many ingredients such as \textit{egg white}, \textit{icing sugar}, \textit{granulated sugar}, and so on. Given Example~\ref{ex_gib_1}, obtained by turning the words of Example~\ref{ex_real_1} into gibberish, a model capable of \textit{generalization} should retrieve the analogous concepts as hypernyms or meronyms (i.e., \textit{twiglomptoroa} is a subclass of \textit{becsverdecoroal} etc.). However, it was not explicitly verified if LLMs would do so, or more broadly, if they are able to generalize taxonomic axioms over text (e.g. textual patterns indicating that $A$ is a subclass of $B$) rather than learn the concepts themselves during the pre-training process.

This paper presents comprehensive experiments to study the generalizability and domain adaptability of LLMs, with perspective of ontology learning. By synthesizing three new domain corpora from the Open English WordNet, and creating a gibberish counterpart each, we assess the performance of LLMs on two main tasks: relation extraction and taxonomy discovery. We conduct the experiments on popular LLMs, including closed- and open-source, off-the-shelf and fine-tuned, and evaluate them on both in-domain and across-domain setups. The novel contributions of this paper are as follows:
\begin{itemize}
    \item We create three synthetic datasets as parallel corpora from the Open English WordNet, by turning domain-specific terms into gibberish.
    \item We conduct experiments to simulate the adaptability and generalizability of LLMs in unseen domains, with or without backpropagation.
    \item We provide empirical evidence that there is a limit to the generalization capability of off-the-shelf LLMs, with OL tasks that leverage lexical semantics such as relation extraction and taxonomy discovery.
    \item We show that in-domain fine-tuning improves in-domain task-specific performance, and that the improvements are transferable to new domains.
\end{itemize}

\section{Related Work}

\textbf{Ontology Learning with LLMs.} 
OL is the (semi)automatic acquisition of \mbox{T-Box} and/or A-Box data from various data sources. In the context of this work, we look at OL from unstructured text or semi-structured data such as Knowledge Graphs paired with textual descriptions. More generally, recent studies show that LLMs can be leveraged to perform ontology related tasks, such as OM (Ontology Matching) or OL. For instance, Norouzi et al.~\cite{norouzi2023conversational} use a naive approach that uses ChatGPT for ontology alignment, by providing the entire source and target ontologies. Mateiu et al.~\cite{mateiu2023ontology} use a fine-tuned GPT-3 to convert natural language into OWL Functional Syntax for ontology enrichment. Hertling et al.~\cite{sven2023olala} use few-shot prompting to enhance the performance of open-source LLMs on OM tasks.

%cite OLALA and other works related to ontology matching with LLMs

In LLMs4OL~\cite{giglou2023llms4ol}, the authors argue that with sufficient formulation, all tasks pertinent to OL fall within one of the three categories: A) Term Typing (determining a generalized type for a lexical term), B) Taxonomy Discovery (determining the hierarchy between a pair of concepts), C) Non-Taxonomic Relation Extraction (finding non-hierarchic relations between concepts). This task paradigm allows them to evaluate LLMs in OL with a zero-shot prompting method. In the work, the authors show that although LLMs may not be suitable for OL as is, they may still be helpful when effectively fine-tuned for ontology construction.

SPIRES~\cite{spires} is a successful application of LLMs to populate ontologies. It leverages Zero-Shot Learning to extract relations between concepts in textual corpora, then grounds the concepts using other existing ontologies in the target domain (e.g. FoodOn or Wikidata). However, in a domain-specific setting, it is not granted that public ontologies of the domain exist and are of high quality.

Moskvoretskii et al.~\cite{Moskvoretskii2024TaxoLLaMAWM} use LLaMa~\cite{touvron2023llama} fine-tuned on WordNet, to perform OL tasks such as taxonomy discovery, taxonomy enrichment, taxonomy construction and lexical entailment. Specifically, they provide further empirical evidence that fine-tuning LLMs for taxonomy discovery on domains drastically increases their performance, making them suitable for the task. The method was tested on real domains such as the food, music and medical domains. In comparison, our work seeks to establish whether or not domain adaptation would hold in arbitrary domains, where the terminology is unknown to the model.

% Our work uses a different approach, but we reach the same conclusion.
%\vspace*{0.2cm}
\noindent \textbf{In-Context Learning with LLMs.}
It was previously observed and verified that LLMs can learn from a few in-context examples in the form of demonstration. In fact, to better answer a given query, a LLM can leverage previous examples to estimate the distribution of input-output pairs. This emergent behavior of LLMs \cite{wei2022emergent} has become a successful learning paradigm because it no longer requires the expensive optimization of model parameters. With respect to our paper, four particular works pertaining to In-Context Learning (ICL) are of high interest. Firstly, Chain-of-Thought prompting~\cite{wei2023chainofthought} forces the model to generate intermediate steps before returning an output, which was shown to elicit reasoning in very large language models and improve their symbolic reasoning performance. Secondly, Min et al.~\cite{min2022rethinking} show that ground-truth labels do not significantly hurt the performance of LLMs on downstream tasks, suggesting that models implicitly learn input-label mappings from the language modelling objective alone. Thirdly, symbol tuning~\cite{wei2023symbol}, the process of fine-tuning by replacing original labels to semantically unrelated ones, was shown to improve the in-context learning performance of very large LMs, and effectively override prior semantic knowledge. Finally, Wei et al.~\cite{wei2023larger} show that``smaller'' LLMs greatly suffer from semantically unrelated labels in comparison to larger ones, heavily implying that they overly rely on semantic priors of targets instead of effectively reasoning over them. The contributions of these works justify our necessity to verify the in-context adaptation of LLMs to domains by extending this verification to semantically unrelated inputs (e.g. gibberish input-label mapings).

% Chain-of-Thought to elicit reasoning
% Larger language models do in-context learning differently
% Finetuned Language Models Are Zero-Shot Learners
% Symbol tuning improves in-context learning in language models
%\vspace*{0.2cm}
\noindent \textbf{Domain Adaptation with LLMs.}
Few works of interest deal with the performance of Large Language Models given domain-specific training corpora.
Wan et al.~\cite{wan2023reformulating} propose a domain adaptation framework for very large language models (GPT-4~\cite{openai2024gpt4}) to address the scarcity of Chinese legal domain texts, using an adapt-retrieve-revise process. With a smaller LLM trained on in-domain corpora, the authors generate a draft answer to retrieve evidence from an external knowledge base, both of which are given to GPT-4 to generate a final answer. Gururangan et al.~\cite{gururangan-etal-2020-dont} show that domain-adaptive pretraining (DAPT), i.e. continued pre-training on domain-specific text, allows one to adapt a language model to a domain at reduced cost. However, Cheng et al.~\cite{cheng2024adapting} argue that while DAPT may improve specific task performance after fine-tuning, it overall hurts the ability of LLMs to perform question answering. Instead, they propose AdaptLLM, a scalable approach that aims to train a LLM on reading comprehension texts created from the raw domain-specific corpora. Finally, Shin et al.~\cite{shin-etal-2022-effect} show that the in-context learning ability of a LLM heavily depends on the corpus sources, and may emerge by combining multiple corpora, but the domain relevance of the corpus may not be indicative of the few-shot performance of the model.

\section{Approach}

\begin{figure}[t]
    \centering
    \includegraphics[width=\textwidth]{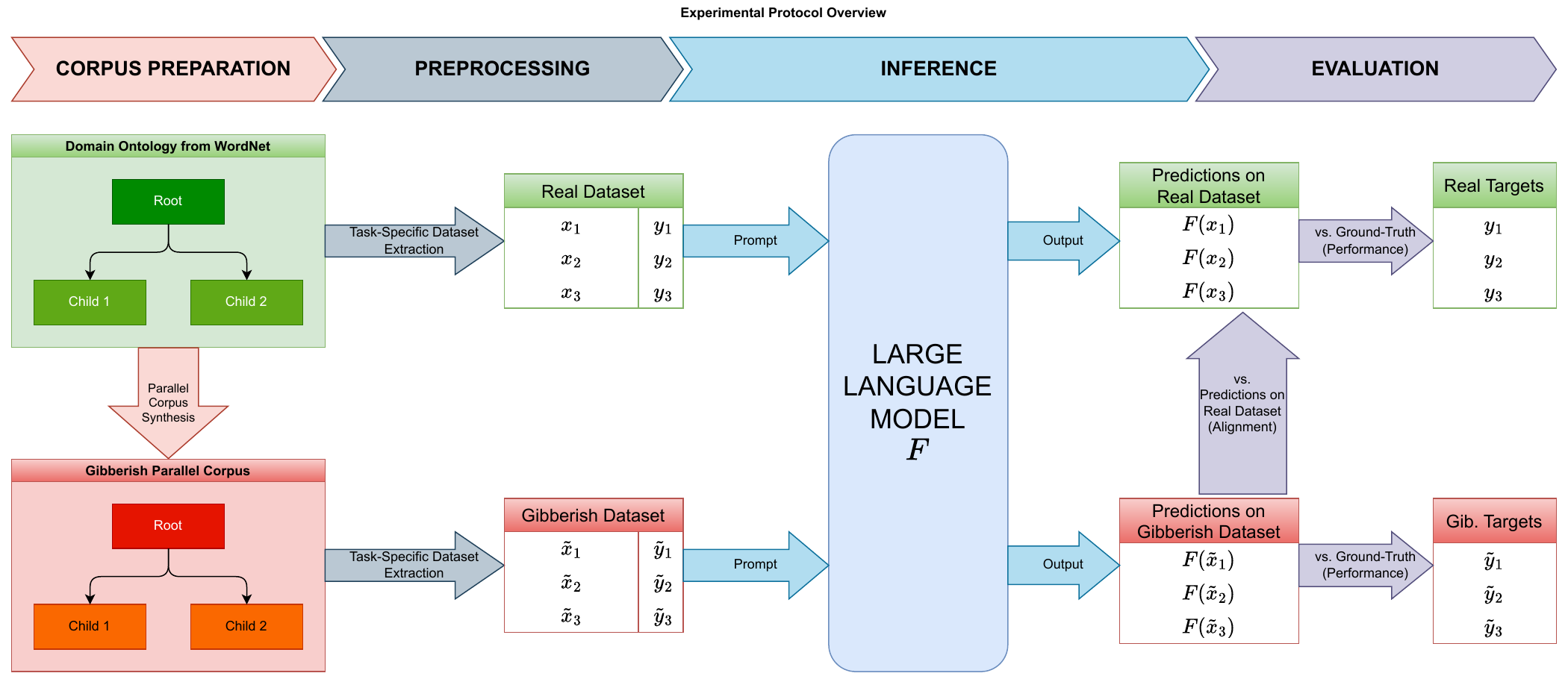}
    \caption{Overview of the off-the-shelf evaluation pipeline. The dataset extraction and the prompt depend on the task to perform.}
    \label{overview}
\end{figure}

Figure~\ref{overview} illustrates an overview of the pipeline we employ to test the \textit{adaptability} and \textit{generalizability} of off-the-shelf LLMs (pretrained, and as is), which includes two main steps: corpus preparation and LLMs evaluation.

% TODO: More details about the accessibility of the code and reproducibility of the experiment

In particular, we use the English WordNet (2023 Edition)~\cite{oewn} to generate a parallel corpus of terms and definitions in the form of gibberish, which serve as our reference domain-specific setting. The process begins by choosing root concepts (for instance, sweets and desserts) in the WordNet taxonomy, then explores the graph through hyponymy, derivation and other (e.g. topic) relationships across concepts (with a set maximal exploration depth). The explored concepts form a domain in the real WordNet (e.g. sweets domain) that can be used to create a parallel corpus by propagating gibberish representations and definitions. More details about the algorithm can be found in Section~\ref{gibberish_dataset}.

After obtaining a parallel corpus of a particular domain, we evaluate a LLM on two different tasks: \textit{relation extraction} and \textit{taxonomy discovery}, each on both versions of the corpus. Naturally, since the concepts remain the same up to an input-label mapping, we ideally expect the results to be analogous. More details can be found in Subsection~\ref{evaluation_protocol}. Additionally, we investigate the effect of fine-tuning on the in-domain performance of LLMs, as described in Subsection~\ref{finetuning_protocol}.

\subsection{Parallel Corpus Synthesis}\label{gibberish_dataset}

%This subsection concerns itself with the creation of a parallel corpus from the English WordNet. 
To simulate a domain that is unseen for the LLM, we generate another KG where the domain concepts have gibberish representations and definitions that do not collide (e.g. if ``sugar'' is turned into ``arghl'' then any definition that contains the word ``sugar'' will see it turned into the word ``arghl'' instead). For this purpose, we devise a procedure which includes three steps: concept mining, concept linking and gibberish generation. 
The code can be found online.\footnote{\url{https://github.com/boschresearch/llm-vs-gibberish-ontologies}}

\begin{comment}
To facilitate the next operations in this section, we will introduce the following Datalog rules, meant to offer shorthand notations.

\input{sections/code/rules}
\end{comment}

\subsubsection{Concept Mining.} The concept mining algorithm is a simple Breadth-First Search algorithm, starting from each of the root concepts $C_0$ and only going through user-selected relationships (hypernymy, sense derivation, and concept topic). We set a maximal exploration depth $D$, which is set to $5$ during experiments. The explored concepts form a dataset $\mathcal{D}$.

\subsubsection{Concept Linking.} The next step of this process is to establish the dependencies between concept definitions and other concepts. For example, if ``sugar'' is mentioned in the definition of concept $c$, then we will link all the concepts $c_{sugar}$ which have ``sugar'' as a representation with $c$ by introducing a blank node as follows, where \texttt{\{c\_id\}} denotes the WordNet ID of $c$:

\begin{lstlisting}[caption=Example of linking triples.]
c   sct:definitionWord    _:_sugar_{c_id} .
_:_sugar_{c_id} rdf:value   "sugar"@en .
_:_sugar_{c_id} sct:references  c_sugar .
\end{lstlisting}\label{linking_example}

\subsubsection{Gibberish Generation.} We assume that we have an algorithm $T$ that creates a gibberish representation $T(c)$ from a concept $c$ based on its initial representation, definition and part-of-speech. A concept is fully processed when it has a gibberish definition AND a gibberish representation. A concept is partially processed when it has a gibberish representation (fully processed implies partially processed).

\begin{figure}[t]
    \centering
     \begin{subfigure}[b]{0.45\textwidth}
         \centering
         \includegraphics[width=\textwidth]{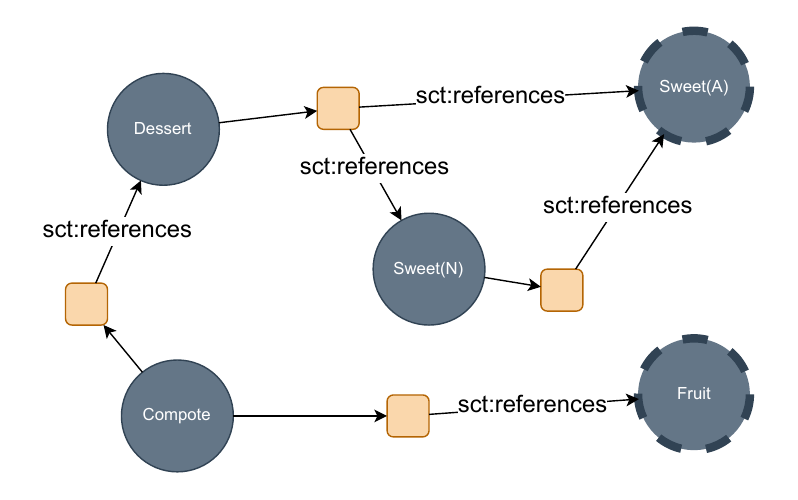}
         \caption{Step 0 : Make a gibberish representation  and (unchanged) definition for concepts with no dependencies.}
         \label{fig:step_0}
     \end{subfigure}
     \hfill
     \begin{subfigure}[b]{0.53\textwidth}
         \centering
         \includegraphics[width=\textwidth]{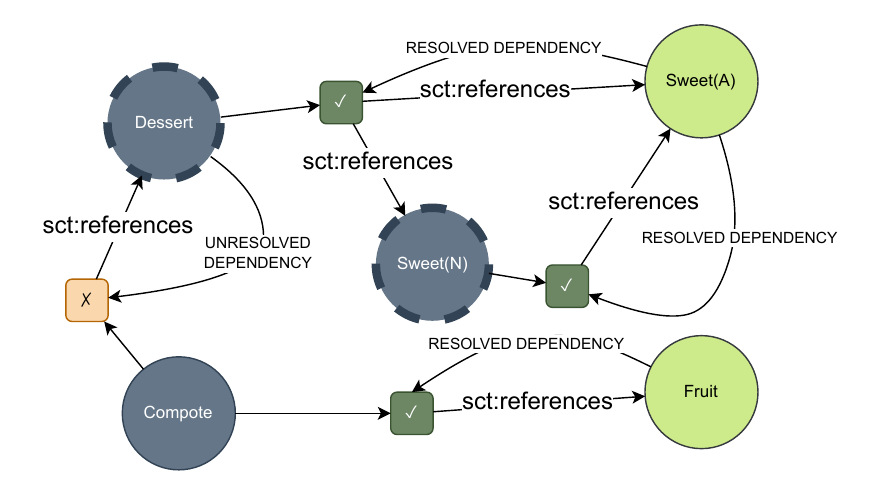}
         \caption{Step $n+1$: Use at least partially processed concepts to resolve dependencies and create new gibberish representations and definitions.}
         \label{fig:step_1}
     \end{subfigure}
    \caption{Algorithm to propagate gibberish representations and definitions. Squares are blank nodes that indicate that the preceding concept has a word that refers to one of the node's successors. Dashed circles represent the concepts eligible to be fully processed. Green circles represent (at least partially) processed concepts.}
    \label{fig:algo_loop}
\end{figure}

Denote $\mathcal{D}_0$ the set of concepts in $\mathcal{D}$ that have no internal dependencies (i.e. for any $c$ in $\mathcal{D}_0$, the definition of $c$ does not refer to any concept in $\mathcal{D}$), as shown in Figure~\ref{fig:step_0}. We create an initial representation $T(c)$ for any $c$ in $\mathcal{D}_0$, and give them a gibberish definition identical to their original one. We will additionally add the homonyms of the previously processed concepts in $\mathcal{D}_0$, with no gibberish representation. Moreover, set $\mathcal{D}_{-1} = \emptyset$. We then repeat the following loop until we have fully processed all the concepts in $\mathcal{D}$, as illustrated in Figure~\ref{fig:step_1}.

Suppose we have obtained $\mathcal{D}_{n-1}$ and $\mathcal{D}_{n-2}$. If $| \mathcal{D}_{n-1} | > | \mathcal{D}_{n-2} |$, we proceed as follows: obtain concepts $c$ that are not fully labeled and which dependencies can all be resolved (i.e. for each dependency \texttt{c sct:definitionWord x}, there exists $c' \in \mathcal{D}_{n-1}$, or $c' = c$ such that \texttt{x sct:references c'}). We create gibberish representations $T(c)$ if there is not one already, we resolve the dependencies using the gibberish representations in $\mathcal{D}_{n-1}$ to make a definition for $c$. All homonyms of processed concepts are also partially processed, and all the partially processed concepts are added to $\mathcal{D}_{n-1}$ to form $\mathcal{D}_{n}$.

If $| \mathcal{D}_{n-1} | = | \mathcal{D}_{n-2} |$, we sample a random concept $c$ of $\mathcal{D}$ that is not partially processed, assign a gibberish representation $T(c)$ to it and add $c$ to $\mathcal{D}_{n-1}$ to form $\mathcal{D}_{n}$. Note that for $i \geq 0$, $\mathcal{D}_i$ does not exclusively contain concepts in $\mathcal{D}$, but $\bar{\mathcal{D}}_i = \mathcal{D}_i \cap \mathcal{D}$ does.

\begin{comment}

\begin{proposition}
    The described algorithm will terminate.
\end{proposition}
\begin{proof}
    The algorithm terminates when every concept in $\mathcal{D}$ is fully processed.
    \begin{itemize}
        \item By design of the loop, either the number of fully processed concepts increases, either the number of partially processed concepts increases.
        \item The number of partially processed concepts is greater than or equal to the number of fully processed concepts at any iteration.
        \item When all concepts in $\mathcal{D}$ are at least partially processed, all definition dependencies can be resolved.
    \end{itemize}
\end{proof}

\end{comment}

This pipeline yields a set of concepts $\mathcal{D}$, where each concept $c \in \mathcal{D}$ has a gibberish representation $T(c)$ as well as a gibberish definition consistent with the internal dependencies in $\mathcal{D}$.

\cameraready{
\subsubsection{Example.} Consider the example depicted in Figure~\ref{fig:algo_loop}. In the first step, Sweet (adjective) and Fruit do not reference any other nodes in their definitions: they can be fully processed. In the second step, since Dessert refers to both Sweet (noun) and Sweet (adjective) and the latter was processed, it is eligible to be processed next. Moreover, since Sweet (adjective) is a homonym of Sweet (noun), it can also be processed. However, since Compote refers to both Dessert and Fruit, it may not be processed yet, since Dessert has not been processed yet. It will be processed in the following step.
}

\subsection{Off-the-Shelf Evaluation Methodology}\label{evaluation_protocol}

For each dataset $\mathcal{D}$, we perform an evaluation by comparing the performance of off-the-shelf LLMs in two different lexical semantic tasks: relation extraction and taxonomy discovery; and compare the outputs coming from two cases: on the original dataset, and its gibberish counterpart.

\subsubsection{Relation Extraction.} Given a query concept in $\mathcal{D}$, its lexical senses (i.e. written forms) and its definition, the goal of this task is to extract all relations between the concepts mentioned in the definition, including the query concept. To remain in the frame of ontology learning, we focus only on hypernymy and holonymy relationships. For instance, in Example~\ref{ex_real_1}, the extracted relations should be as follows: macaron is a subclass of confection, egg white is a part of macaron, icing sugar is a part of macaron, etc. Likewise, given the gibberish definition in \ref{ex_gib_1} instead, we expect the same relations to be extracted, only with the concept names replaced with their gibberish counterpart. In order to retrieve a prediction, the LLM is prompted to output triples in the form: $(A, r, B)$ where $r$ is either \texttt{is a subclass of} or \texttt{is a part of}.

\subsubsection{Taxonomy Discovery.} Given two concepts A and B in $\mathcal{D}$, with their lexical senses and definitions, the goal of this task is to determine whether or not A is a subclass of B. Likewise, we expect that turning the lexical senses and definitions of A and B into gibberish will not change the outcome. WordNet hypernymy relations are used as a ground-truth: the predictions on the real (resp. gibberish counterpart of the) dataset are compared with the real (resp. gibberish counterpart of the) ground-truth. Indirect hypernymy relations (obtained with the transitive closure of the relation \texttt{gwn:hypernym}) are also used. Negative examples are produced by corrupting the hypernym once or twice per query (hyponym, subclass of, ?). This classification task can be evaluated with F1-score: a drop in performance should indicate that a LLM relies on the lexical senses to infer taxonomical relationships rather than textual semantic information.

\subsubsection{Prompting.} For each task/model configuration, we follow general guidelines for prompting a LLM. (1) The return is in JSON format, (2) Chain-of-Thought (CoT)~\cite{wei2023chainofthought} can be used if it improves the performance of the model, (3) One or few exemplar(s) can be used, with example(s) outside of the dataset, if it improves the performance of the model. In the \textbf{relation extraction} task, for a concept $C$, its written form $F_C$, its part-of-speech $P_C$, and its definition $D_C$, we construct a prompt $p(C)$ as shown in Listing~\ref{lst:extraction_template}. In the \textbf{taxonomy discovery} task, for $A, B$ two concepts, $F_A, F_B$ their written forms, $D_A, D_B$ their definition, we construct a prompt $p(A,B)$ as shown in Listing~\ref{lst:discovery_template}. Prompt templates can be found \href{https://github.com/boschresearch/llm-vs-gibberish-ontologies}{here}.
% TODO: ADD LINK TO PROMPT TEMPLATES

\begin{lstlisting}[caption={Relation extraction generic prompt $p(C)$.}, label={lst:extraction_template}]
{FORMAT INSTRUCTIONS (Task and return format)}
{EXAMPLE(S) (zero, one or few exemplars)}

Concept: {F_C}
Part-of-speech: {P_C}
Definition: {D_C}
\end{lstlisting}

\begin{lstlisting}[caption={Taxonomy discovery generic prompt $p(A,B)$.}, label={lst:discovery_template}]
{FORMAT INSTRUCTIONS (Task and return format)}
{EXAMPLE(S) (zero, one or few exemplars)}

Concept A: {F_A}
Definition: {D_A}

Concept B: {F_B}
Definition: {D_B}
\end{lstlisting}

\subsection{Fine-tuning Experiment}\label{finetuning_protocol}

While it was previously verified that fine-tuning improves OL performance in existing domains~\cite{Moskvoretskii2024TaxoLLaMAWM}, it is not clear if this statement still holds for arbitrary domains and unknown vocabularies where reasoning is required. To verify this question, after evaluating the zero/one/few-shot performance of LLMs in the gibberish domains, we propose to assess the effect of fine-tuning models on the inference performance for a specific task. 

\subsubsection{Data split.} For each dataset $\mathcal{D}$, we fine-tune a LLM for taxonomy discovery on a train split of hypernymy relations. Half of the concepts in the dataset, and their hypernymy relations, are used for training. The inverse relations are used as negatives (if A is a subclass of B, then B is not a subclass of A), alongside some randomly sampled negatives. The remaining relations are used for testing.

\subsubsection{Prompting.} We train LLMs on instructive datasets with the prefix prompt in Listing~\ref{prompts_train}, completed using each pair of concepts in the hypernymy dataset.

\begin{lstlisting}[caption={Prefix prompt used for the fine-tuning experiment. The content between curly brackets is formatted for each pair of concepts in the dataset.},label={prompts_train}]
### HUMAN:
Identify whether the statement is true or false. Answer with only one word: 'True' or 'False'.

CONCEPT A: {term_a} ({pos_a})
Definition: {definition_a}

CONCEPT B: {term_b} ({pos_b})
Definition: {definition_b}

Statement: '{term_a}' is a subclass of '{term_b}'.
### ASSISTANT:
\end{lstlisting}

\section{Experiments}

In this section, we explore the off-the-shelf evaluation proposed in Subsection~\ref{evaluation_protocol}, and the fine-tuning experiments described in Subsection~\ref{finetuning_protocol}.

\subsection{Experimental setup}

\subsubsection{Datasets.} To assess the performance variation with domain specificity, we generate three synthetic domain-specific datasets as parallel corpora from the Open English WordNet~\cite{oewn}, using the methodology described in Section~\ref{gibberish_dataset}. They are:
\begin{itemize}
    \item Sweets: A collection of concepts related to sweets, desserts, sweet food or sugar. In this dataset, hypernyms are frequent, and concepts are relatively well constructed from their hypernyms.
    \item Football: A collection of concepts related to football. This dataset, created by browsing co-topic concepts, includes less taxonomic relationships, but has its own terminology and jargon.
    \item Music: A collection of concepts related to musical instruments. It is the largest of the three.
\end{itemize}
Table~\ref{tab:statistics} shows, for each dataset, the number of concepts, the number of hypernymy relationships, the exploration depth and the root concepts. The translator $T$ used to generate gibberish representations of concepts is available online.\footnote{\url{https://github.com/htmai-880/gibberify}}.

\begin{table}[t]
\centering
\caption{Dataset statistics. The `Hypernyms' column indicates the number of retrieved hypernymy relationships within the dataset. The `Depth' column indicates the exploration depth of the algorithm.}
\resizebox{\textwidth}{!}{
\begin{tabular}{ccccc}
\toprule
\multirow{2}{*}{$\mathcal{D}$} & \multirow{2}{*}{\textbf{Concepts}} & \multirow{2}{*}{\textbf{Hypernyms}} & \multirow{2}{*}{\textbf{Depth}} & \multirow{2}{*}{\textbf{Root Concepts}} \\
 & & & & \\
\midrule
\textbf{WN-sweets} & 244 & 418 & 5 & sweet (n), sweet (a), sugar\\
\textbf{WN-football} & 937 & 1401 & 5 & football, team, offensive (a), defensive (a) \\
\textbf{WN-music} & 1366 & 2497 & 5 & musical instrument \\
\bottomrule
\end{tabular}
}
\label{tab:statistics}
\end{table}

\subsubsection{Models.}
While this study is not comprehensive of all existing LLMs, our goal is to show a particular trend between them. In the off-the-shelf evaluation, we evaluate the following popular LLMs, with their number of parameters between parentheses: GPT-3.5~\cite{brown2020language} (174B), GPT-4~\cite{openai2024gpt4} ($\geq$ 1T), Falcon-40B~\cite{falcon40b} (40B), LLaMa2-13B~\cite{touvron2023llama} (13B), and Zephyr-7B-$\beta$~\cite{tunstall2023zephyr} (7B). The former two, accessed with a paid subscription, are closed-source, whereas the latter three are open-source. In the fine-tuning experiment, we consider Zephyr-7B-$\beta$~\cite{tunstall2023zephyr} and Falcon-7B~\cite{falcon40b}, which are both open-source. For the paid subscription models, We limit our budget to 15\euro{} per dataset.

\subsubsection{Specificities.} We henceforth consider the three following evaluation methods:
\begin{itemize}
    \item Ground-truth (GT)(\texttt{en}) vs \texttt{en}  we compare the answers on the original English dataset against the ground-truth
    \item Ground-truth (GT)(\texttt{gib}) vs \texttt{gib}: we compare the answers on the gibberish dataset against the gibberish ground-truth
    \item \texttt{en} vs \texttt{gib}: using the answers on the original dataset as a ground-truth, we evaluate the consistency of the predictions on the gibberish dataset, regardless of their correctness.
\end{itemize}
In the relation extraction (GT(X) vs X), due to the scarcely annotated holonymy relationships in WordNet, we focus on hypernymy relationships to compute metrics. Moreover, a model prediction is processed by taking into account the inferred hypernymy relationships, using the transitive property of the relation. For instance, if a model outputs the triples (vanilla pudding, is a subclass of, custard-like pudding), and (custard-like pudding, is a subclass of, pudding), we count the triple (vanilla pudding, is a subclass of, pudding) as effectively predicted by the model. In the taxonomy discovery task, predictions that are neither ``True" or ``False" are ignored. Thus, macro-averaged F1-scores may not be between the macro-averaged precisions and the macro-averaged recalls.

\subsection{Off-the-Shelf Evaluation}

\subsubsection{Metrics.} For each model/task configuration, we compute metrics in three settings: GT(\texttt{en}) vs \texttt{en}, GT(\texttt{gib}) vs \texttt{gib}, and \texttt{en} vs \texttt{gib}. The goal is to see if the predictions with gibberish terms align with the predictions with the real terms. We compute the following metrics: precision, recall and F1-score.

\begin{table}[t]
    \centering
    \caption{Results on the relation extraction task in the GT(X) vs X setting, where X is either \texttt{en} or \texttt{gib}. Only hypernymy relations are considered.}
    \resizebox{\textwidth}{!}{
    \begin{tabular}{cp{1.cm}p{1cm}p{1cm}p{1cm}p{1cm}p{1cm}p{1cm}p{1cm}p{1cm}p{1cm}}
    \toprule
         \multirow{2}{*}{\textbf{Model}} & \multirow{2}{*}{\textbf{X}} & \multicolumn{3}{c}{\textbf{WN-sweets}} & \multicolumn{3}{c}{\textbf{WN-football}} & \multicolumn{3}{c}{\textbf{WN-music}}  \\
         \cline{3-11}
         & & Pre. & Rec. & F1 & Pre. & Rec. & F1 & Pre. & Rec. & F1\\
    \midrule
    \multirow{2}{*}{GPT-3.5} & \texttt{en} & 0.478 & 0.150 & 0.228 & 0.383 & 0.056 & 0.097 & 0.397 & 0.060 & 0.104 \\
         & \texttt{gib} & 0.336 & 0.069 & 0.115 & 0.371 & 0.035 & 0.065
         & 0.307 & 0.029 & 0.053 \\
         \midrule
    \multirow{2}{*}{GPT-4} & \texttt{en} & 0.583 & 0.160 & 0.251 & - & - & - & - & - & - \\
         & \texttt{gib} & 0.530 & 0.129 & 0.207 & - & - & - & - & - & - \\
         \midrule
    \multirow{2}{*}{Falcon-40B} & \texttt{en} & 0.573 & 0.151 & 0.238 & 0.489 & 0.067 & 0.118 & 0.529 & 0.065 & 0.116 \\
         & \texttt{gib} & 0.330 & 0.080 & 0.128 & 0.382 & 0.050 & 0.088 & 0.341 & 0.042 & 0.074 \\
         \midrule
    \multirow{2}{*}{LLaMa2-13B} & \texttt{en} & 0.536 & 0.141 & 0.223 & 0.423 & 0.035 & 0.065 & 0.434 & 0.030 & 0.056 \\
         & \texttt{gib} & 0.365 & 0.085 & 0.138 & 0.341 & 0.018 & 0.035 & 0.296 & 0.014 & 0.026 \\
         \midrule
    \multirow{2}{*}{Zephyr-7B-$\beta$} & \texttt{en} & 0.441 & 0.158 & 0.233 & 0.399 & 0.067 & 0.115 & 0.374 & 0.063 & 0.108 \\
         & \texttt{gib} & 0.243 & 0.095 & 0.137 & 0.313 & 0.052 & 0.089 & 0.261 & 0.044 & 0.075 \\
    \bottomrule
    \end{tabular}
    }
    \label{tab:gt_rel_extraction_results}
\end{table}

\begin{table}[t]
    \centering
        \caption{Results (macro-average) on the taxonomy discovery task in the GT(X) vs X setting, where X is either \texttt{en} or \texttt{gib}.}
    \resizebox{\textwidth}{!}{
    \begin{tabular}{cp{1.cm}p{1cm}p{1cm}p{1cm}p{1cm}p{1cm}p{1cm}p{1cm}p{1cm}p{1cm}}
    \toprule
         \multirow{2}{*}{\textbf{Model}} & \multirow{2}{*}{\textbf{X}} & \multicolumn{3}{c}{\textbf{WN-sweets}} & \multicolumn{3}{c}{\textbf{WN-football}} & \multicolumn{3}{c}{\textbf{WN-music}}  \\
         \cline{3-11}
         & & Pre. & Rec. & F1 & Pre. & Rec. & F1 & Pre. & Rec. & F1\\
    \midrule
    \multirow{2}{*}{GPT-3.5} & \texttt{en} & 0.944 & 0.937 & 0.940 & 0.758 & 0.701 & 0.648 & 0.829 & 0.858 & 0.818 \\
         & \texttt{gib} & 0.783 & 0.539 & 0.446 & 0.640 & 0.505 & 0.333 & 0.687 & 0.537 & 0.361 \\
         \midrule
    \multirow{2}{*}{GPT-4} & \texttt{en} & 0.949 & 0.943 & 0.945 & - & - & - & - & - & - \\
         & \texttt{gib} & 0.591 & 0.576 & 0.566 & - & - & - & - & - & - \\
         \midrule
    \multirow{2}{*}{Falcon-40B} & \texttt{en} & 0.775 & 0.658 & 0.598 & 0.800 & 0.648 & 0.637 & 0.787 & 0.620 & 0.613 \\
         & \texttt{gib} & 0.591 & 0.575 & 0.574 & 0.483 & 0.475 & 0.478 & 0.541 & 0.535 & 0.480 \\
         \midrule
    \multirow{2}{*}{LLaMa2-13B} & \texttt{en} & 0.819 & 0.772 & 0.750 & 0.808 & 0.811 & 0.809 & 0.785 & 0.800 & 0.790 \\
         & \texttt{gib} & 0.450 & 0.447 & 0.444 & 0.533 & 0.531 & 0.504 & 0.576 & 0.556 & 0.465 \\
         \midrule
    \multirow{2}{*}{Zephyr-7B-$\beta$} & \texttt{en} & 0.899 & 0.897 & 0.898 & 0.813 & 0.751 & 0.759 & 0.821 & 0.762 & 0.778 \\
         & \texttt{gib} & 0.691 & 0.634 & 0.621 & 0.500 & 0.500 & 0.469 & 0.530 & 0.524 & 0.523 \\
    \bottomrule
    \end{tabular}
    }
    \label{tab:gt_tax_disc_results}
\end{table}

\begin{table}[t]
    \centering
        \caption{Alignment results (macro-average) on the relation extraction task in the \texttt{en} vs \texttt{gib} setting. Both hypernymy and holonymy relations are considered.}
    \resizebox{\textwidth}{!}{
    \begin{tabular}{cp{1cm}p{1cm}p{1cm}p{1cm}p{1cm}p{1cm}p{1cm}p{1cm}p{1cm}}
    \toprule
         \multirow{2}{*}{\textbf{Model}} & \multicolumn{3}{c}{\textbf{WN-sweets}} & \multicolumn{3}{c}{\textbf{WN-football}} & \multicolumn{3}{c}{\textbf{WN-music}}  \\
         \cline{2-10}
         & Pre. & Rec. & F1 & Pre. & Rec. & F1 & Pre. & Rec. & F1\\
    \midrule
    \multirow{1}{*}{GPT-3.5}  & 0.371 & 0.304 & 0.334 & 0.263 & 0.175 & 0.210 & 0.207 & 0.138 & 0.166 \\
    \multirow{1}{*}{GPT-4}  & 0.504 & 0.527 & 0.515 & - & - & - & - & - & - \\
    \multirow{1}{*}{Falcon-40B}  & 0.310 & 0.303 & 0.306 & 0.236 & 0.238 & 0.237 & 0.225 & 0.224 & 0.225 \\
    \multirow{1}{*}{LLaMa2-13B}  & 0.347 & 0.340 & 0.344 & 0.386 & 0.225 & 0.284 & 0.374 & 0.215 & 0.273 \\
    \multirow{1}{*}{Zephyr-7B-$\beta$}  & 0.214 & 0.229 & 0.221 & 0.192 & 0.180 & 0.185 & 0.148 & 0.142 & 0.145 \\
    \bottomrule
    \end{tabular}
    }
    \label{tab:alignment_extraction}
\end{table}

\begin{table}[t]
    \centering
        \caption{Alignment results (macro-average) on the taxonomy discovery task in the \texttt{en} vs \texttt{gib} setting.}
    \resizebox{\textwidth}{!}{
    \begin{tabular}{cp{1cm}p{1cm}p{1cm}p{1cm}p{1cm}p{1cm}p{1cm}p{1cm}p{1cm}}
    \toprule
         \multirow{2}{*}{\textbf{Model}} & \multicolumn{3}{c}{\textbf{WN-sweets}} & \multicolumn{3}{c}{\textbf{WN-football}} & \multicolumn{3}{c}{\textbf{WN-music}}  \\
         \cline{2-10}
         & Pre. & Rec. & F1 & Pre. & Rec. & F1 & Pre. & Rec. & F1\\
    \midrule
    \multirow{1}{*}{GPT-3.5}  & 0.789 & 0.541 & 0.465 & 0.700 & 0.517 & 0.488 & 0.738 & 0.544 & 0.459 \\
    \multirow{1}{*}{GPT-4}  & 0.611 & 0.594 & 0.590 & - & - & - & - & - & - \\
    \multirow{1}{*}{Falcon-40B}  & 0.541 & 0.557 & 0.412 & 0.502 & 0.495 & 0.443 & 0.529 & 0.562 & 0.352 \\
    \multirow{1}{*}{LLaMa2-13B}  & 0.565 & 0.570 & 0.556 & 0.618 & 0.610 & 0.586 & 0.641 & 0.599 & 0.535 \\
    \multirow{1}{*}{Zephyr-7B-$\beta$}  & 0.818 & 0.727 & 0.716 & 0.557 & 0.544 & 0.545 & 0.570 & 0.572 & 0.570 \\
    \bottomrule
    \end{tabular}
    }
    \label{tab:alignment_discovery}
\end{table}

\subsubsection{Results.} We first examine the results with respect the to ground-truths.
It is important to mention that the English WordNet is scarcely annotated in terms of hypernymy and holonymy relationships. Consider the following example:%Example~\ref{toffee_apple}.
\begin{mdframed}
\begin{example}[toffee apple]
    an apple that is covered with a candy-like substance (usually caramelized sugar).
\label{toffee_apple}
\end{example}
\end{mdframed}

%In Example~\ref{toffee_apple}
In this example, the definition obviously implies that a \textit{toffee apple} is an \textit{apple}, but WordNet only considers \textit{sweet, confection} to be valid hypernyms. The incompleteness of WordNet explains why the observed performances are so low, but because our goal is a relative comparison of performances on real corpora and their gibberish counterpart, rather than absolute scores, it is not \cameraready{critical} to have a high-quality ground-truth.

In both tasks, which results are reported in Table~\ref{tab:gt_rel_extraction_results} and Table~\ref{tab:gt_tax_disc_results}, a common trend occurs across all LLMs and in all synthetic domains: a significant performance decrease is observed when replacing real terms with gibberish. Although GPT-4 (tested on WN-sweets only because of the slowness and the costs) performs best on gibberish corpora, it suffers from the same performance drop.

While the performance on relation extraction is generally low across all LLMs with the real datasets, which we mainly attribute to the poor quality of the annotations, it is even lower with the gibberish datasets. In each case, the macro F1-score is practically halved, e.g. in WN-sweets, GPT-3.5 drops from 0.228 to 0.115, Falcon-40B drops from 0.238 to 0.128, LLaMa2-13B drops from 0.223 to 0.138, Zephyr-7B-$\beta$ drops from 0.233 to 0.137. Note that the recall is low in comparison to the precision, which indicates that a LLM tends to overlook indicators that two concepts are hierarchically related. This observation aligns with the conclusion of LLMs4OL~\cite{giglou2023llms4ol}, according to which off-the-shelf LLMs are not sufficiently suitable for OL tasks, particularly in the case of relation extraction.

In the taxonomy discovery task, the performance of LLMs also plummets when using the gibberish dataset instead of the real one. The LLMs are relatively good at identifying whether two real concepts are hierarchically related (e.g. F1-score up to 0.940 on the WN-sweets dataset by GPT-3.5), but suffer from a large performance drop when confronted to unknown words which share the same semantic relations (e.g. the F1-score of GPT-3.5 drops from 0.940 to 0.446 when using the gibberish counterpart of WN-sweets; the same behavior is observed across all datasets and LLMs). We interpret this drop as the fact that LLMs are significantly better at leveraging semantic priors (i.e. lexical senses known from pre-training), to deduce that Concept A is a subclass of Concept B.

Although the general performance drop is expected because the tested LLM would then be dealing with words it has never seen during its training, we furthermore observe that the prediction alignment is very low. In spite of analogous concepts sharing the same semantic relations with each other in the parallel corpora, the model is unable to produce analogous outputs from analogous inputs, as shown by the low F1-scores in Tables~\ref{tab:alignment_extraction} and~\ref{tab:alignment_discovery}. This is evidence that as is, LLMs do not \textit{reason} over semantic relationships.

Our interpretation is that the attention mechanism heavily relies on the lexical sense and the frame of a token, instead of leveraging the semantic relationships that hold between tokens. In other words, the ``reasoning'' abilities of LLMs for Ontology Learning are mostly limited to entities and concepts that the models have already been trained on, i.e. prior semantics. However, such a quality is critical for Ontology Learning in arbitrary domains, where hypernymy relationships must be retrieved for unknown concepts, or new concepts that share the same lexical form as some existing word (for instance, if domain-specific jargon employs existing words with different meanings).

This trend seems to hold true in both tasks, which confirms the fact that off-the-shelf LLMs are currently not suited for OL tasks on arbitrary domains.

\subsection{Fine-Tuning Evaluation}

\begin{table}[t]
\centering
\caption{Hypernym train-test split for the fine-tuning experiment.}
\begin{tabular}{ccccc}
\toprule
\multirow{2}{*}{$\mathcal{D}$} & \multicolumn{2}{c}{\textbf{Train}} & \multicolumn{2}{c}{\textbf{Test}}\\
& Positives & Negatives & Positives & Negatives\\
\midrule
\textbf{WN-sweets} & 189 & 393 & 229 & 284\\
\textbf{WN-football} & 674 & 1043 & 727 & 567 \\
\textbf{WN-music} & 1367 & 1882 & 1130 & 851 \\
\bottomrule
\end{tabular}
\label{tab:split}
\end{table}

\subsubsection{Training details.} Table~\ref{tab:split} shows the number of hypernym pairs used for the fine-tuning experiment. We use instruction tuning specifically tailored towards taxonomy discovery. In the prefix prompt documented in Listing~\ref{prompts_train}, given two concepts $A$ and $B$, the model must return ``True" if $A$ is a subclass of $B$ and ``False" otherwise. We train the model on 20 epochs, with a batch size of $4$ and a learning rate of $3 \cdot 10^{-6} $. For efficient training, we quantize the model on 4 bits and use the LoRA~\cite{hu2021lora} method with parameters $\alpha = 256$ and $r = 1024$.

\subsubsection{Metrics.} Similarly to the first experiment, we use Precision, Recall and F1-score to evaluate the performance of the model on the testing set.

\begin{table}[t]
    \centering
    \caption{Taxonomy discovery results with fine-tuning in the GT(X) vs X setting, where X is either \texttt{en} or \texttt{gib}. Evaluation is performed before and after fine-tuning.}
    \resizebox{\textwidth}{!}{
    \begin{tabular}{cp{1.cm}p{1cm}p{1cm}p{1cm}p{1cm}p{1cm}p{1cm}p{1cm}p{1cm}p{1cm}}
    \toprule
         \multirow{2}{*}{\textbf{Model (X)}} & \multirow{2}{*}{\textbf{When}} & \multicolumn{3}{c}{\textbf{WN-sweets}} & \multicolumn{3}{c}{\textbf{WN-football}} & \multicolumn{3}{c}{\textbf{WN-music}}  \\
         \cline{3-11}
         & & Pre. & Rec. & F1 & Pre. & Rec. & F1 & Pre. & Rec. & F1\\
    \midrule
    \multirow{2}{*}{Falcon-7B (\texttt{en})} & Before & 0.564 & 0.507 & 0.388 & 0.561 & 0.504 & 0.325 & 0.631 & 0.509 & 0.327 \\
         & After &  0.923 & 0.902 & 0.907 & 0.867 & 0.871 & 0.868 & 0.874 & 0.881 & 0.874 \\
         \midrule
    \multirow{2}{*}{Falcon-7B (\texttt{gib})} & Before & 0.275& 0.491 & 0.352 & 0.594 & 0.501 & 0.309 & 0.314 & 0.498 & 0.300 \\
         & After & 0.725 & 0.663 & 0.655 & 0.685 & 0.687 & 0.679 & 0.708 & 0.703 & 0.683 \\
         \midrule
    \multirow{2}{*}{Zephyr-7B-$\beta$ (\texttt{en})} & Before & 0.898 & 0.845 & 0.853 & 0.772 & 0.679 & 0.618 & 0.783 & 0.722 & 0.674 \\
         & After & 0.905 & 0.906 & 0.897 & 0.940 & 0.939 & 0.939 & 0.941 & 0.939 & 0.940 \\
         \midrule
    \multirow{2}{*}{Zephyr-7B-$\beta$ (\texttt{gib})} & Before & 0.746 & 0.572 & 0.500 & 0.740 & 0.621 & 0.535 & 0.723 & 0.589 & 0.479 \\
         & After & 0.840 & 0.816 & 0.796 & 0.859 & 0.810 & 0.817 & 0.839 & 0.846 & 0.839 \\
         \midrule
    \end{tabular}
    }
    \label{tab:finetuning_exp_res}
\end{table}

\begin{figure}[t]
    \centering
    \includegraphics[width=\textwidth]{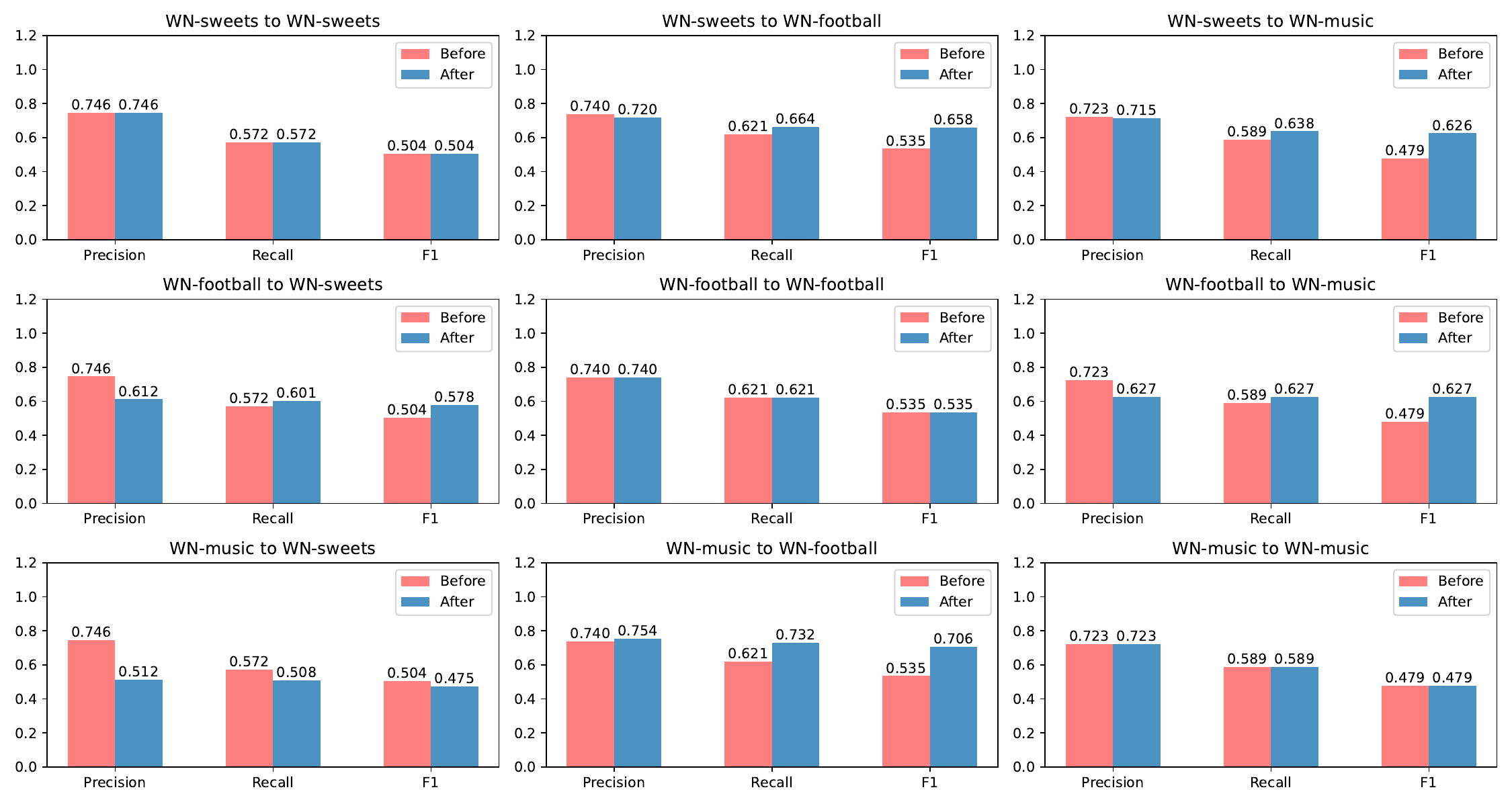}
    \caption{Domain transfer experiment results with Zephyr-7B-$\beta$. Only gibberish corpora are selected for both training and testing.}
    \label{fig:transfer_results}
    % \vspace{-0.4cm}
\end{figure}

\subsubsection{Results.} The results of the fine-tuning experiment are reported in Table~\ref{tab:finetuning_exp_res}. Two general observations can be made about the experiment.
Firstly, we notice that fine-tuning drastically improves the task-specific performance of the LLM regardless the domain and corpus version. For instance, while Falcon-7B seems to be initially worse-performing than Zephyr-7B-$\beta$ overall, the F1-score improves up to almost threefold (real WN-music dataset, from 0.327 to 0.874). Although not surprising for real corpora, an improvement on gibberish corpora is nontrivial: the LLMs show signs of adaptation on gibberish corpora with improved performance. Secondly, while the performance of a model on a gibberish corpus increases after fine-tuning, it never matches the performance of the same model fine-tuned on the real counterpart of the corpus. It is worth noting that this is a limitation of adaptation solely due to the reliance on prior semantics, since a corpus and its gibberish counterpart only differ in their input-label mapping.

\subsubsection{Transfer Learning.} In spite of the two previous observations, we can hypothesize that the improvement in performance on the gibberish corpora may indicate signs of reasoning and generalization on unseen domains, given that gibberish words are most likely not contained in the vocabulary of LLMs. To elaborate this claim, we propose another experiment: we use Zephyr-7B-$\beta$, trained for taxonomy discovery in a domain, and test it on another domain. By only using gibberish corpora, we ensure that most of the domain-specific terminology is anonymized, preventing the LLM from effectively using prior semantics. Results are reported in Figure~\ref{fig:transfer_results}. We observe the following. Firstly, the F1-scores generally tend to increase after transfer, with the exception of the WN-music to WN-sweets case. The increase in F1-score is substantial, from 14\% (WN-football to WN-sweets), up to 32\% (WN-music to WN-football). This result is quite promising, as it points towards the possibility of OL on arbitrary domains with LLMs with effective pre-training. Secondly, the precision tends to drop in favor of the recall, indicating that fine-tuning makes the LLM more sensitive to syntactic clues of hypernymy relations at the cost of making them less precise. Due to the fact that both the training and the testing domains are made of gibberish words, we attribute the performance improvements of the LLM (with respect to its base version) to emerging reasoning capabilities: the fine-tuning LLM becomes more capable of abstraction and of focusing on semantic relationships between concepts, rather than the concepts themselves.

\section{Conclusion}

We have explored and tested the limits of adaptability and generalizability of LLMs, and observed that LLMs do not adapt well to arbitrary domains. By creating gibberish datasets based on real data and real domains from WordNet, and using LLMs to perform ontology learning tasks on these data, it is realized that LLMs are unable to consistently retrieve the same taxonomic relationships between analogous concepts, which highlights their clear reliance on priorly learned semantics, lexical senses, and the frame of the tokens. However, we notice that after fine-tuning on gibberish data, LLMs improve at discovering hierarchies, both on the domain they were trained on and other arbitrary domains. We attribute this improvement to the emergence of reasoning with lexical semantics. Our work serves as cautionary advice for the community that LLMs do not adapt to arbitrary domains, and we hope that it can inspire future work to leverage reasoning with LLMs for Ontology Learning.

\paragraph*{Supplemental Material Statement:} Generated datasets (real and gibberish for all domains), source code for generating the synthetic datasets from the Open English WordNet, and for fine-tuning or evaluating LLMs on OL tasks are available online.\footnote{\url {https://github.com/boschresearch/llm-vs-gibberish-ontologies}} 

\subsubsection{\ackname}
The work was partially supported by EU project: enRichMyData (HORIZON-CL4-2021-DATA-01 - GA 101070284).

\bibliographystyle{splncs04}
\bibliography{bibliography}
\end{document}